# A Novel Statistical Method Based on Dynamic Models for Classification

[1]Lerong Li and Momiao Xiong[1,*]

[1]Human Genetics Center, Division of Biostatistics, the University of Texas School of Public Health, Houston, TX 77030.

Running Title: Dynamic models for classification

[*]Address for correspondence and reprints: Dr. Momiao Xiong, Human Genetics Center, the University of Texas Health Science Center at Houston, P.O. Box 20186, Houston, Texas 77225, (Phone): 713-500-9894, (Fax): 713-500-0900, E-mail: Momiao.Xiong@uth.tmc.edu.

**Summary**

Realizations of stochastic process are often observed temporal data or functional data. There are growing interests in classification of dynamic or functional data. The basic feature of functional data is that the functional data have infinite dimensions and are highly correlated. An essential issue for classifying dynamic and functional data is how to effectively reduce their dimension and explore dynamic feature. However, few statistical methods for dynamic data classification have directly used rich dynamic features of the data. We propose to use second order ordinary differential equation (ODE) to model dynamic process and principal differential analysis to estimate constant or time-varying parameters in the ODE. We examine differential dynamic properties of the dynamic system across different conditions including stability and transient-response, which determine how the dynamic systems maintain their functions and performance under a broad range of random internal and external perturbations. We use the parameters in the ODE as features for classifiers. As a proof of principle, the proposed methods are applied to classifying normal and abnormal QRS complexes in the electrocardiogram (ECG) data analysis, which is of great clinical values in diagnosis of cardiovascular diseases. We show that the ODE-based classification methods in QRS complex classification outperform the currently widely used neural networks with Fourier expansion coefficients of the functional data as their features. We expect that the dynamic model-based classification methods may open a new avenue for functional data classification.

**Introduction**

Realizations of stochastic process are often observed temporal data or functional data. There are growing interests in classification of functional data. For example, classification of electrocardiogram (ECG) is of great clinical value (Ye, Kumar and Coimbra, 2012). QRS complex that is also referred to the cardiac beats is an essential part of the ECG signals. It provides valuable information on the biological process of the heart. Automatic ECG analysis requires the accurate detection of abnormal QRS complex and hence demands on efficient statistical methods for classification of functional data.

A basic feature of the functional data is that the functional data have infinite dimensions and are highly correlated, which makes estimation of the covariance matrix difficult. There is increasing evidence that standard multivariate statistical analysis often fails with functional data. A common solution is to choose a finite dimensional basis and project the functional curve onto this basis (James and Hastie, 2001). Then the resulting basis coefficients form a finite dimensional representation. Therefore, the widely used statistical methods for functional data classification consist of two steps. At the first step the functional data are projected into a finite-dimensional space of basis functions or eigenfunction (Luo L, 2013; Ramsay and Silverman, 2005). At the second step, all the methods for classification of multivariate data can be used to classify the coefficients in the expansions by linear discriminant analysis (LDA) (James and Hastie, 2001), k-nearest neighbor method (Biau, Bunea and Wegkamp, 2005) ,support vector machine (Rossi and Villa, 2006) and by wavelet and more general discrimination rules (Berlinet, Biau and Rouviere, 2008).

To more efficiently use functional dependent information, highlight some features of the data and improve the classification accuracy, functional data are often transformed in several ways (Alonso, Casado and Romo, 2012).The data are, in general, centered and normalized (Rossi and Villa, 2006). The most important data transformation is the derivation. A linear combination of the original functional curves and their different order of derivatives are then projected onto basis functions or eigenfunctions (Alonso et al., 2012).Incorporating dynamic information (derivative information) into feature will capture the dependence characteristics of the functional data and hence improve classification.

Although incorporating functional derivative into features increase potential to improve classification accuracy, taking functional derivatives as features will substantially increase the number of features. We hope that we will not increase the number of features while we employ dynamic information in classifying functional curves. Using the parameters in differential equations that model the dynamics of continuously changing process such as QRS complex in ECG data as features can serve this purpose. Differential equations are widely used powerful tools to describe dynamic systems in many physical, chemical and biological processes (Liang and Wu, 2008; Poyton et al., 2006). Parameters in differential equations capture fundamental behaviors of the real dynamic processes and are consistent with the available data. Only a few parameters are needed to capture essential dynamic features of the systems. Therefore, we propose to use parameters in differential equations that model dynamic process such as QRS complex as features for classification.

Estimation of parameters in ordinary differential equations is a challenge problem. There are two approaches. First approach is a two-stage method (Cao, Huang and Wu, 2012; Chen and Wu, 2008; Liang and Wu, 2008; Wu, Xue and Kumar, 2012). In the first stage, regression spline is used to estimate true functional curve and its derivatives. In the second stage, parameters are estimated by minimizing the square of errors between the estimated derivatives by regression spline and the predicted derivatives from the model. Second approach is principal differential analysis method (Poyton et al., 2006).It consists of two steps. The first step is to fit the dynamic or functional data by basis function expansion such as B-spline. The second step is to use the fitted curves (e.g., resulting empirical spline curves) and their derivatives to transform the differential equations to algebraic equations and use the least square methods to estimate the parameters.

It is increasingly recognized that biological systems as a whole are not just the sum of their components but, rather, ever-changing, complex, interacted and dynamic systems over time in response to internal events and environmental stimuli (Assmus et al., 2006).The phenomena of the entire organisms are not only determined by steady-state characteristics of the biological systems, but also determined by inherent dynamic properties of biological systems which are largely governed by the internal structure of the system.  Dynamic properties include stability and transient-response, which determine how the systems maintain their functions and performance under a broad range of random internal and external perturbations and their responses to changes in environments. Similar to allele frequency differences or differential gene expressions between cases and controls, we can also observe the differential dynamic properties of the biological systems across different types of tissues and conditions.  Dynamic properties are correlated with the health status of individuals and are of central importance to comprehensively understanding of human biological systems and classifying dynamic data or functional data. However, to our knowledge, very few statistical methods for functional data classification directly explore rich dynamic features in the functional data that are generated by dynamic systems.

The purpose of this paper is to develop the ODE-based methods for classifying dynamic or functional data. To accomplish this, we first extend the principal differential analysis for the parameter estimation from the first order ODE (Poyton et al., 2006) to the second order ODE with both constant and time-varying parameters and then use the ECG data as examples to examine the accuracy of the ODE for fitting the dynamic data. We apply mathematical methods and computational algorithms from engineering and control theory (Ogata, 1997) to perform dynamic analysis for both normal and abnormal QRS complex in the ECG data set. We investigate the differential dynamic properties between the normal and abnormal QRS complexes and reveal the relationship between the dynamic properties of the dynamic systems and their phenotypes. We use the vector support machine (SVM) and the parameters in the ODE as its input features to classify functional or dynamic data. To evaluate its performance, finally, the proposed methods are applied to the real ECG data from MIT-BIH arrhythmia database to classify QRS complex. Our preliminary results show that the ODE-based method for classifying QRS complex outperforms the widely used neural networks (NN) that use the Fourier expansion coefficients of the QRS complex curves as their input features. A program for implementing the developed ODE-based methods for dynamic data classification can be downloaded from our website http://www.sph.uth.tmc.edu/hgc/faculty/xiong/index.htm and http://www.bioconductor.org/.

**2. Differential Equation for Extracting Features of Dynamic Process**

**2.1. Differential equation with constant and time-varying parameters for modeling a dynamic system**

We assume that $x(t)$ is a state variable in a dynamic system which can be modeled by the following second-order ordinary differential equation (ODE) with constant parameter:

$$L(x(t)) = \frac{d^2 x(t)}{dt^2} + w_1 \frac{dx(t)}{dt} + w_0 x(t) = 0 \qquad \text{or} \qquad (1a)$$

the second-order differential equation with time-varying parameters:

$$L(x(t)) = \frac{d^2 x(t)}{dt^2} + w_1(t) \frac{dx(t)}{dt} + w_0(t) x(t) = 0, \qquad (1b)$$

where $w_1$ ($w_1(t)$) and $w_0$ ($w_0(t)$) are weighting coefficients or parameters in the ODE. The state $x(t)$ is hidden. Its observations $y(t)$ often have measurement errors:

$$y(t) = x(t) + e(t), \qquad (2)$$

where $e(t)$ is measurement error at the time $t$.

**2.2. Principal differential analysis for estimation of parameters in differential equations.**

The estimators of the parameters in the ODE can be obtained by principal differential analysis (Poyton et al., 2006). The purpose of parameter estimation is to attempt to determine the appropriate parameter values that make the errors between the predicted response values and the measured data as small as possible. The predicted response values can be obtained by solving ODE for modeling the dynamic system. One way to solve ODE is to first expand the function $x(t)$ in terms of basis functions. Let $x_i(t)$ be the state variable at time $t$ of the $i$-th sample satisfying ODE (1a) or 1b) and $y_i(t)$ be its observation ($i = 1,...,n$). Then, $x_i(t)$ cab be expanded as

$$x_i(t) = \sum_{j=1}^{K} c_{ij}\phi_j(t) = C_i^T \phi(t), \qquad (3)$$

where $C_i = [c_{i1},...,c_{iK}]^T$ and $\phi(t) = [\phi_1(t),...,\phi_K(t)]^T$.

Similarly, the parameters $w_1(t)$ and $w_0(t)$ can be expanded as

$$w_1(t) = \sum_{j=1}^{K} h_{1j}\phi_j(t) = h_1^T \phi(t) \text{ and}$$

$$w_0(t) = \sum_{j=1}^{K} h_{0j}\phi_j(t) = h_0^T \phi(t), \qquad (4)$$

where $h_1 = \begin{bmatrix} h_{11} \\ \vdots \\ h_{1K} \end{bmatrix}$ and $h_0 = \begin{bmatrix} h_{01} \\ \vdots \\ h_{0K} \end{bmatrix}$.

Let

$$\psi(t) = \frac{d^2\phi}{dt^2} + \frac{d\phi}{dt}\phi^T(t)h_1 + \phi(t)\phi^T(t)h_0$$
$$= \frac{d^2\phi}{dt^2} + G(t)h,$$

and

$$J_{\phi h} = \int_T \psi(t)\psi^T(t)dt,$$

where $G(t) = \begin{bmatrix} \dfrac{d\phi}{dt}\phi^T(t) & \phi(t)\phi^T(t) \end{bmatrix}, h = \begin{bmatrix} h_1 \\ h_0 \end{bmatrix}$.

The differential operator is given by $L(x_i(t)) = C_i^T \psi(t)$. The penalty term is defined as

$$\lambda \int_T L(x_i(t)) L^T(x_i(t)) dt = \lambda C_i^T J_{\phi h} C_i.  \qquad (5)$$

We estimate the state function $x(t)$ from the observation data $y(t)$ by minimizing the following objective function which consists of the sum of the squared errors between the observations and the states and the penalty terms:

$$\sum_{i=1}^{n} \left\{ \sum_{j=1}^{T} [y_i(t_j) - x_i(t_j)]^2 + \lambda \int_T L(x_i(t)) L^T(x_i(t)) dt \right\}$$
$$= \sum_{i=1}^{n} \left\{ \sum_{j=1}^{T} [y_i(t_j) - \phi^T(t_j) C_i]^2 + \lambda C_i^T J_{\phi h} C_i \right\} \qquad (6)$$

Let

$$Y_i = \begin{bmatrix} y_i(t_1) \\ \vdots \\ y_i(t_T) \end{bmatrix}, \tilde{\phi} = \begin{bmatrix} \phi^T(t_1) \\ \vdots \\ \phi^T(t_T) \end{bmatrix}, Y = \begin{bmatrix} Y_1 \\ \vdots \\ Y_n \end{bmatrix}, \Phi = \begin{bmatrix} \tilde{\phi} & 0 & \cdots & 0 \\ 0 & \tilde{\phi} & & 0 \\ \vdots & \vdots & \ddots & 0 \\ 0 & 0 & \cdots & \tilde{\phi} \end{bmatrix}, J = \begin{bmatrix} J_{\phi h} & 0 & \cdots & 0 \\ 0 & J_{\phi h} & \cdots & 0 \\ \vdots & \vdots & \ddots & \vdots \\ 0 & 0 & \cdots & J_{\phi h} \end{bmatrix}, C = \begin{bmatrix} C_1 \\ \vdots \\ C_n \end{bmatrix}.$$

The problem (6) can then be reduced to

$$\min_C (Y - \Phi C)^T (Y - \Phi C) + \lambda C^T J C. \qquad (7)$$

The Least square estimators of the expansion coefficients are then given by

$$C = (\Phi^T \Phi + \lambda J)^{-1} \Phi^T Y. \qquad (8)$$

Next we estimate the parameters in the ODE. The parameters in the ODE can be estimated by minimizing the following least squares objective function:

$$\min_h \text{SSE}_p = \int_T L^T(X(t)) L(X(t)) dt, \qquad (9)$$

where $L(X(t)) = [L(x_1(t)),...,L(x_n(t))]^T$. Since $L(x_i(t)) = C_i^T \psi(t)$, the $L(X(t))$ can be expressed in terms of the estimated expansion coefficients as

$$L(X(t)) = \hat{C}\Psi(t).$$

Therefore, problem (9) can be reduced as

$$\min_{h} \ \text{SSE}_p = \int_T \psi^T(t)\hat{C}^T\hat{C}\psi(t)dt, \qquad (10)$$

where the matrix $\hat{C}$ is estimated and hence fixed in the minimization problem (10). Setting partial derivative of $\text{SSE}_p$ to be zero:

$$\frac{\partial \text{SSE}_p}{\partial h} = \int_T G^T(t)\hat{C}^T\hat{C}[\frac{d^2\phi(t)}{dt^2} + G(t)h]dt = 0. \qquad (11)$$

Solving equation (11) for $h$, we obtain

$$h = -[\int_T G^T(t)\hat{C}^T\hat{C}G(t)dt]^{-1}\int_T G^T(t)\hat{C}^T\hat{C}\frac{d^2\phi}{dt^2}dt. \qquad (12)$$

In summary, we iteratively determine the expansion coefficients for fixed parameters in the ODE by equation (8) and estimate the parameters in the ODE for fixed expansion coefficients by equation (12) until convergence.

## 3. Results

### 3.1. Dynamic model for QRS complex

Automatic ECG analysis has wide applications in heartbeat classification. Manually beat-by-beat examination is very time-consuming and tedious in the recognition of classes of consecutive heartbeats (Ye et al., 2012). Automatic ECG analysis is able to monitor cardiac activity for timely detection of abnormal heart conditions and diagnosis of cardiac arrhythmias. The heart beats are also often referred to QRS complex shapes. Three letters Q, R and S stand for the three main phases of a cardiac cycle (Ravier et al., 2007). Numerous methods for heartbeat classifications have been developed in the past several decades. These methods use a number of features such as Hermite coefficients (Lagerholm et al., 2000), wavelet features (Ince, Kiranyaz and Gabbouj, 2009), waveform shape features (de Chazal and Reilly, 2006) and RR interval information (Ye et al., 2012). However, heart beat is a dynamic process. These most currently used features for heart beat classification do not fully employ dynamic features of QRS complex. We propose to use ODE to model QRS complex which can fully capture dynamic features of QRS complex. Let $x(t)$ be the magnitude of the signal in QRS complexes at the time point $t$ and $y(t)$ be its observed signal in ECG curves. The second order ODE for modeling the dynamics of the signals in the QRS complex is given by

$$\frac{d^2 x(t)}{dt^2} + w_1(t)\frac{dx(t)}{dt} + w_0(t)x(t) = 0$$
$$y(t) = x(t) + \varepsilon,$$
(13)

where errors $\varepsilon$ are independently distributed as a normal distribution with mean of zero and variance $\sigma^2$.

**3.2. B-spline fitting and parameter estimation in ODE**

We apply the proposed methods to estimate the parameters in equation (13) for modeling QRS complex. We first evaluate how well the B-spline to fit the signal $x(t)$ in the QRS complex. The data were from MIT-BIH arrhythmia database, which contained 48 half-hours excerpts of two channel ambulatory ECG signals (lead MLII and lead V5; only lead MLII used) from 47 subjects. The ECG signals were band-pass filtered at 0.1-100 Hz and sampled at 360 Hz per channel with 11-bit resolution over a 10 mV range (http://www.physionet.org/). Two or more cardiologists independently annotated each record; disagreements were resolved to obtain the computer-readable reference annotations for each beat included with the database. And there was an annotation file associated with each record to provide the reference annotations like QRS location and heartbeat categories which were used to extract QRS segments and true labels for each heartbeat respectively.

ECG data were measured and recorded by skin electrodes. These data were easily contaminated by various types of noises due to power-line interference, muscle contraction and electrode movements, and electromyography (Sun et al., 2012). To remove unnecessary noise and improve the signal-to-noise ratio, the ECG waveform is band-pass filtered at 5-12 Hz (low-band and high band pass filtered) to remove baseline wander and power-line interference. And the filtered ECG signal is used for dynamical model fitting (Pan and Tompkins, 1985).

Figures 1 and 2 plotted the observed signal, the fitted curve by the cubic B-spline with a uniform knot spacing of 0.012 second, and the trajectories of solutions to the second ODE with constant parameters and time-varying parameters for normal and abnormal QRS complex, respectively. The tuning parameter was chosen by cross-validation at each iteration of the smoothing. Both Figures showed that the cubic B-spline and the second ODE with the time-varying parameters can approximate the observed signals very well in both normal and abnormal complexes. However, the trajectories of the solutions to the second ODE with constant parameters cannot approximate the observed signals very well and are only similar to the shapes of the curves of the observed signals.

### 3.3. Stability and transient-response analysis

Although the second ODE with the constant parameters can only approximate the shape of the curves of the observed signals, it can characterize the dynamic behaviors of the QRS complex remarkably well. Dynamic properties include stability and transient-response, which determine how the systems maintain their functions and performance under a broad range of random internal and external perturbations and their responses to changes in environments.

The most important dynamic property of biological systems is concerned with stability. A dynamic system is called stable if their state variables return to, or towards their original states or equilibrium states following internal and external perturbations (Kremling and Saez-Rodriguez, 2007). The stability of the system is a property of the system itself. One of the methods for assessing the stability of the dynamic system is to analyze eigenvalues of the eigenequation of the high-order ODE which models the linear dynamic systems. If the real parts of all eigenvalues of the dynamic system are negative then the system is stable and if the real parts of all eigenvalues of the dynamic system are positive then the system is unstable. The coefficients of the second ODE for the normal QRS complex in Figure 1 are $w_1 = 2.598$ and $w_0 = 9394.2$ and its eigenvalues are $\lambda_1 = -1.30 + 96.9i$ and $\lambda_2 = -1.30 - 96.9i$. The coefficients of the second ODE for the abnormal QRS complex in Figure 2 are $w_1 = -6.97$ and $w_0 = 4535.9$ and its eigenvalues are $\lambda_1 = 3.48 + 67.26i$ and $\lambda_2 = 3.48 - 67.26i$. We clearly observed that the real parts of all eigenvalues of the second ODE for the normal QRS complex were negative, but the real parts of all eigenvalues for the abnormal QRS complex were positive. This showed that the normal QRS complex was stable, but the abnormal complexes were unstable. The dynamic system in heart will remain at steady state until occurrence of external perturbation. Depending on dynamic behavior of the system after perturbation of environments, the steady-states of the system are either stable (the system returns to the initial state) or unstable (the system leaves the initial equilibrium state). For any practical purpose, the dynamic QRS complex must be stable. Unstable dynamic system underlying abnormal QRS complex will lead to the irregular activities in heart or even to the heart failure.

The dynamic behavior of the cardiac underlying QRS complex is encoded in the temporal evolution of its states. Response of a biological system to perturbation of internal and external stimuli has two parts: the transient and the steady state response. The process generated in going from the initial state to the final state in the response to the perturbation of the internal and external stimuli is called transient response. Steady-state response studies the system behavior at infinite time. Transient-response analysis of the cardiac system can be used to quantify their dynamics. It can reveal how fast the dynamic system in heart responds to perturbation of environments and how accurately the system in heart can finally achieve the desired steady-state values. It can also be used to study damped vibration behavior and stability of the cardiac dynamic system.

The transient response of a dynamic system depends on the input signals. Different signals will cause different responses. There are numerous types of signals in practice. For the convenience of comparison, we consider two types of signals: (i) unit-step signal and (ii) unit-impulse signal.

The transfer function of the response of the cardiac dynamic system underlying the QRS complex to unit-step and unit-impulse input signals are given by $Y(s) = \frac{G(s)}{S}$ and $Y(s) = G(s)$ respectively, where $G(s)$ is the transfer function of the dynamic system. The transient-response analysis of the dynamic system can be performed by inverse Laplace transformation. We performed the transient-response analysis with MATLAB (Ogata, 1997). Figures 3 and 4 showed the unit-step response curves of cardiac system underlying normal and abnormal QRS complexes, respectively. They showed that the unit-step response cardiac system underlying normal and abnormal QRS complexes were substantially different. Although we observed signal oscillations in both normal and abnormal QRS complex, while the signals from the normal QRS complex after the unit-step perturbation quickly reach the steady states, the signals from the abnormal QRS complex to respond to the perturbation were highly oscillated and would never reach the steady-state values. This phenomenon suggested that dynamic responses of abnormal QRS complex to environmental stimuli were irregular. Figures 5 and 6 showed the unit-impulse response curves of cardiac system underlying normal and abnormal QRS complexes, respectively. We observed similar patterns to that of the unit-step response of QRS complexes.

**3.4. Classification of signals from dynamic systems**

Previous studies strongly demonstrated that the second-order ODE can capture dynamic features of QRS complexes very well. Next we evaluate the performance of using the parameters in the second-order ODE to classify QRS complexes. We used three measures to evaluate the classification performance: sensitivity, specificity and accuracy. Sensitivity is defined as the percentage of the true abnormal signals correctly classified as abnormal. Specificity is defined as the percentage of the true normal signals correctly classified as normal. The classification accuracy is defined as the percentage of the correctly classified normal and abnormal signals. We used 19 free ECG records of the MIT/BIH database (Ravier et al., 2007). The parameters in the second-order ODE were used as features of the QRS complexes and input to the support vector machine (SVM) (Chang and Lin, 2011) to classify QRS complexes. We also compare our results with the neural network classifier that was used in the paper (Ravier et al., 2007). For the time-varying parameters in the second-order ODE, we used functional principal component scores (Ramsay and Silverman, 2005) of the time-varying parameter functions $w_0(t)$ and $w_1(t)$ in the second-order ODE as features of the SVM. Specifically, the QRS complex signals were expanded in terms of Fourier series. Their Fourier expansion coefficients of the QRS complexes were used as features for a multilayer neural network (NN) with 16 input nodes, 4 neurons in the hidden layer, and one output neuron. For both ODEs with constant and time-varying parameters we also used the height of the R point and width of the ARS complex as their features.

The results of classification were summarized in Table 1 where ODE is referred to as the ODE with constant parameters, ODET, the ODE with time-varying parameters and NN, neural networks. Several features emerge from the results in Table 1. First, in most cases, the ODE with constant parameters has the highest accuracy. Second, it is surprisingly observed that although the trajectory of the solution to ODE with time-varying parameters approximates the observed signal curve of QRS complex is much better than that of the ODE with constant parameters, its performance for classifying QRS complex is not as good as that of the ODE with constant parameters in many cases. Third, when the numbers of the sampled normal and abnormal QRS complexes were highly imbalanced (records: 51, 100, 101, 202) either sensitivity or specificity of all three classifiers were clearly not accurate because the training samples did not provide enough information on distinguishing normal QRS complex from abnormal QRS complex. In general, when the number of sampled normal QRS complexes was substantially larger than that of abnormal QRS complexes the specificity of classification was high. In contrast, if the number of abnormal complexes is larger than that of normal QRS complexes we can obtain high sensitivity. Fourth, when the numbers of both normal and abnormal QRS complexes were balanced, both sensitivity and specificity can reach high values.

**4. Discussion**

In this paper we have studied how the ODE can be used to classify dynamic systems and functional data. We have revealed that the ODE can capture dynamic features of the biological processes such as ECG and perform classification of dynamic processes very well. To facilitate application of the ODE-based methods for classifying dynamic and functional data we have addressed several issues in developing ODE-based methods for dynamic system classification.

The first issue is whether the ODE can accurately model the rapidly changing dynamic processes such as ECG signals. The accurate parameter estimation is an essential to the success of the ODE for modeling data. We have extended the iterated principal differential analysis for estimating the constant or time-varying parameters from the first-order ODE to the second-order ODE. We showed that the cubic B-spline can fit the noisy and dramatically fluctuating QRS complex signals very well, which resulted in accurate parameter estimates and QRS signal estimates.

Second issue is whether the second-order ODE can capture the dynamic features of the data or not. The stability and transient response are two key features of the dynamic system. The stability of the system is to characterize the ability of the system to return to the equilibrium states after perturbation of the internal and external stimuli. The requirements for stable biological systems are necessary conditions for the normal operations of the organisms. The dynamic behavior of a system is encoded in the temporal evolution of its states. The transient response of the dynamic system to environmental changes characterizes its dynamical process to respond the perturbation of environments. We used the second-order ODE with constant parameters to investigate the stability and transient response of the normal and abnormal complexes. Interestingly, we observe the differential dynamic properties of the QRS complexes across different types of conditions. We found that normal QRS complex is stable, but abnormal QRS complex is instable. We also found that the signals from the normal QRS complex after the perturbation quickly reach the steady states; the signals from the abnormal QRS complex to respond to the perturbation were highly oscillated and would never reach the steady-state values.

Third issue is the ability of the ODE to discriminate the normal dynamic system from abnormal dynamic systems. We observed that the second-order ODE-based methods can accurately classify QRS complex and outperform the classical neural networks with the Fourier expansion coefficients of the QRS complex signals as their input. We also observed that second-order ODE with constant parameters also outperforms the second-order ODE with the time-varying parameters. The second–order ODE with constant parameters only has two parameters. This demonstrated that the ODE for modeling QRS complex can dramatically reduce the dimensions of the functional QRS complex data while retaining the high accuracy for classifying QRS complexes.

Although the preliminary results are appealing, they suffer from several limitations. The first limitation, i.e., the most important limitation is that we cannot automatically detect abnormal QRS complex. We used supervised learning methods to classify QRS complex which require training data. In other words, we need to label a set of QRS complexes by cardiologists as the training samples before we can use the second-order ODE to classify the new QRS complexes. We need to develop the dynamic models to cluster QRS complexes which will lead to the automatic detection of abnormal QRS complexes. This will be very important in clinical diagnosis of irregular heartbeat patients. Second limitation is that the QRS complex is only a part of an ECG signal curve. Therefore, it is much easier to model the QRS complexes than to model the whole ECG curve. We need (1) to develop the high order ODE for modeling the whole ECG curve, (2) to develop efficient statistical methods for estimating the parameters in the high order ODE and (3) to use the parameters in the high order ODE to classify cardiovascular diseases such as myocardial infarction detection.

Our results in this paper are preliminary. Our intension is to stimulate further discussions about how to use dynamic features of the observed functional data for classification. We expect that the ODE will open a new way to classify functional and dynamic data.

**Acknowledgments**

The project described was supported by Grant 1R01HL106034-01 and 1R01GM104411-01 from the National Institutes of Health, and Award Number R25DA026120 from the National Institute on Drug Abuse.

**Web Resources**

A program for implementing the proposed methods can be downloaded from our website http://www.sph.uth.tmc.edu/hgc/faculty/xiong/index.htm and http://www.bioconductor.org/

**References:**

Alonso, A. M., Casado, D., and Romo, J. (2012). Supervised classification for functional data: A weighted distance approach. *Computational Statistics & Data Analysis* **56**, 2334-2346.

Assmus, H. E., Herwig, R., Cho, K. H., and Wolkenhauer, O. (2006). Dynamics of biological systems: role of systems biology in medical research. *Expert review of molecular diagnostics* **6**, 891-902.

Berlinet, A., Biau, G., and Rouviere, L. (2008). Functional supervised classification with wavelets. In *Annales de l'ISUP*.

Biau, G., Bunea, F., and Wegkamp, M. H. (2005). Functional classification in Hilbert spaces. *IEEE Trans. Inf. Theor.* **51**, 2163-2172.

Cao, J., Huang, J. Z., and Wu, H. (2012). Penalized nonlinear least squares estimation of time-varying parameters in ordinary differential equations. *Journal of Computational and Graphical Statistics* **21**, 42-56.

Chang, C.-C., and Lin, C.-J. (2011). LIBSVM: a library for support vector machines. *ACM Transactions on Intelligent Systems and Technology (TIST)* **2**, 27.

Chen, J., and Wu, H. (2008). Efficient local estimation for time-varying coefficients in deterministic dynamic models with applications to HIV-1 dynamics. *Journal of the American Statistical Association* **103**, 369-384.

de Chazal, P., and Reilly, R. B. (2006). A patient-adapting heartbeat classifier using ECG morphology and heartbeat interval features. *Biomedical Engineering, IEEE Transactions on* **53**, 2535-2543.

Ince, T., Kiranyaz, S., and Gabbouj, M. (2009). A generic and robust system for automated patient-specific classification of ECG signals. *Biomedical Engineering, IEEE Transactions on* **56**, 1415-1426.

James, G. M., and Hastie, T. J. (2001). Functional Linear Discriminant Analysis for Irregularly Sampled Curves. *Journal of the Royal Statistical Society. Series B (Statistical Methodology)* **63**, 533-550.

Kremling, A., and Saez-Rodriguez, J. (2007). Systems biology—an engineering perspective. *Journal of biotechnology* **129**, 329-351.

Lagerholm, M., Peterson, C., Braccini, G., Edenbrandt, L., and Sornmo, L. (2000). Clustering ECG complexes using Hermite functions and self-organizing maps. *Biomedical Engineering, IEEE Transactions on* **47**, 838-848.

Liang, H., and Wu, H. (2008). Parameter estimation for differential equation models using a framework of measurement error in regression models. *Journal of the American Statistical Association* **103**.

Luo L, Z. Y. a. X. M. (2013). Penalized functional principal component analysis for sequence-based association studies. *Eur J Hum Genet*, 21.

Ogata, K. (1997). *System Dynamics*, Third Edition edition.

Pan, J., and Tompkins, W. J. (1985). A real-time QRS detection algorithm. *Biomedical Engineering, IEEE Transactions on*, 230-236.

Poyton, A., Varziri, M. S., McAuley, K. B., McLellan, P., and Ramsay, J. O. (2006). Parameter estimation in continuous-time dynamic models using principal differential analysis. *Computers & chemical engineering* **30**, 698-708.

Ramsay, J. J. O., and Silverman, B. W. (2005). *Functional Data Analysis*: Springer Science+Business Media, Incorporated.

Ravier, P., Leclerc, F., Dumez-Viou, C., and Lamarque, G. (2007). Redefining performance evaluation tools for real-time QRS complex classification systems. *Biomedical Engineering, IEEE Transactions on* **54**, 1706-1710.

Rossi, F., and Villa, N. (2006). Support vector machine for functional data classification. *Neurocomputing* **69**, 730-742.

Sun, L., Lu, Y., Yang, K., and Li, S. (2012). ECG Analysis Using Multiple Instance Learning for Myocardial Infarction Detection.

Wu, H., Xue, H., and Kumar, A. (2012). Numerical Discretization-Based Estimation Methods for Ordinary Differential Equation Models via Penalized Spline Smoothing with Applications in Biomedical Research. *Biometrics* **68**, 344-352.

Ye, C., Kumar, B. V., and Coimbra, M. T. (2012). Heartbeat classification using morphological and dynamic features of ECG signals. *IEEE transactions on bio-medical engineering* **59**, 2930-2941.

Table 1. Accuracy of ODE constant and time varying oarameters, and neural networks for classifying QRS complex

| File # | # of Normal QRS | # of Abnormal QRS | Sensitivity | | | Specificity | | | Accuracy | | |
|---|---|---|---|---|---|---|---|---|---|---|---|
| | | | ODE | ODET | NN | ODE | ODET | NN | ODE | ODET | NN |
| 100 | 2233 | 34 | 0.059 | 0.118 | 0.235 | 0.999 | 0.999 | 0.995 | 0.985 | 0.985 | 0.983 |
| 101 | 1851 | 8 | 0.250 | 0.000 | 0.167 | 1.000 | 1.000 | 1.000 | 0.997 | 0.996 | 0.997 |
| 102 | 99 | 2082 | 1.000 | 1.000 | 0.998 | 0.980 | 0.909 | 0.788 | 0.999 | 0.996 | 0.988 |
| 104 | 163 | 2060 | 0.969 | 0.992 | 0.983 | 0.699 | 0.675 | 0.783 | 0.950 | 0.969 | 0.964 |
| 105 | 2484 | 110 | 0.745 | 0.273 | 0.622 | 0.988 | 1.000 | 0.996 | 0.978 | 0.969 | 0.990 |
| 106 | 1500 | 511 | 0.990 | 0.990 | 0.939 | 0.999 | 0.970 | 0.978 | 0.997 | 0.975 | 0.968 |
| 119 | 1539 | 444 | 1.000 | 1.000 | 0.995 | 1.000 | 0.995 | 0.992 | 1.000 | 0.996 | 0.993 |
| 200 | 1736 | 847 | 0.919 | 0.939 | 0.867 | 0.997 | 0.929 | 0.963 | 0.971 | 0.932 | 0.931 |
| 201 | 1604 | 272 | 0.853 | 0.673 | 0.747 | 0.999 | 0.993 | 0.991 | 0.978 | 0.946 | 0.956 |
| 202 | 2044 | 69 | 0.391 | 0.188 | 0.306 | 0.999 | 0.997 | 0.991 | 0.979 | 0.971 | 0.971 |
| 205 | 2564 | 83 | 0.867 | 0.084 | 0.790 | 1.000 | 1.000 | 0.995 | 0.995 | 0.971 | 0.988 |
| 208 | 1573 | 1356 | 0.973 | 0.982 | 0.945 | 0.987 | 0.941 | 0.960 | 0.980 | 0.960 | 0.953 |
| 209 | 2616 | 384 | 0.444 | 0.447 | 0.482 | 0.981 | 0.978 | 0.976 | 0.912 | 0.910 | 0.912 |
| 210 | 2410 | 212 | 0.882 | 0.269 | 0.770 | 0.995 | 0.999 | 0.954 | 0.986 | 0.940 | 0.940 |
| 212 | 919 | 1824 | 0.967 | 0.971 | 0.958 | 0.984 | 0.928 | 0.958 | 0.973 | 0.957 | 0.958 |
| 213 | 2634 | 609 | 0.849 | 0.875 | 0.863 | 0.995 | 0.983 | 0.987 | 0.968 | 0.963 | 0.964 |
| 215 | 2412 | 120 | 0.958 | 0.196 | 0.800 | 0.997 | 0.904 | 0.986 | 0.995 | 0.705 | 0.979 |
| 217 | 242 | 1959 | 0.995 | 0.993 | 0.987 | 0.946 | 0.946 | 0.964 | 0.990 | 0.988 | 0.984 |
| 219 | 1539 | 51 | 0.824 | 0.278 | 0.667 | 1.000 | 0.862 | 0.994 | 0.994 | 0.697 | 0.986 |

**Figure Legend**

**Figure 1**.  Observed signal (in green color), fitted curve (in blue color) by cubic B-spline, the trajectories of solutions to differential equations with constant parameters (in pink color) and time-varying parameters (in red color) in normal QRS complex (153th QRS complex of the 208th individual from the MIT-BIH arrhythmia database).

**Figure 2.**  Observed signal (in green color), fitted curve (in blue color) by cubic B-spline, the trajectories of solutions to differential equations with constant parameters (in pink color) and time-varying parameters (in red color) in abnormal QRS complex (50th QRS complex of the 208th individual from the MIT-BIH arrhythmia database).

**Figure 3**.  Unit-step response curve of cardiac system underlying normal QRS complex.

**Figure 4**.  Unit-step response curve of cardiac system underlying abnormal QRS complex.

**Figure 5**.  Unit-impulse response curve of cardiac system underlying normal QRS complex.

**Figure 6**.  Unit-impulse response curve of cardiac system underlying abnormal QRS complex.

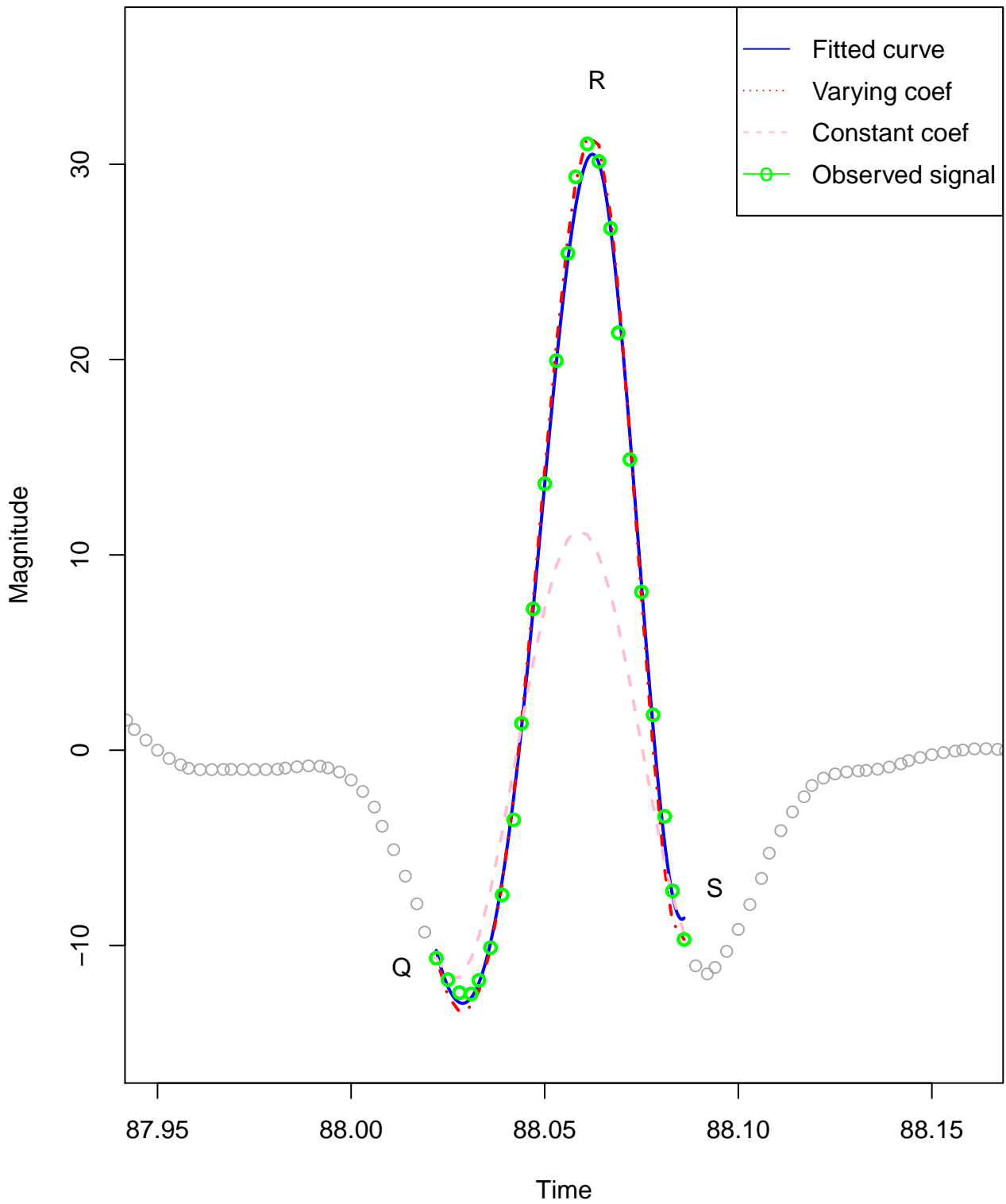

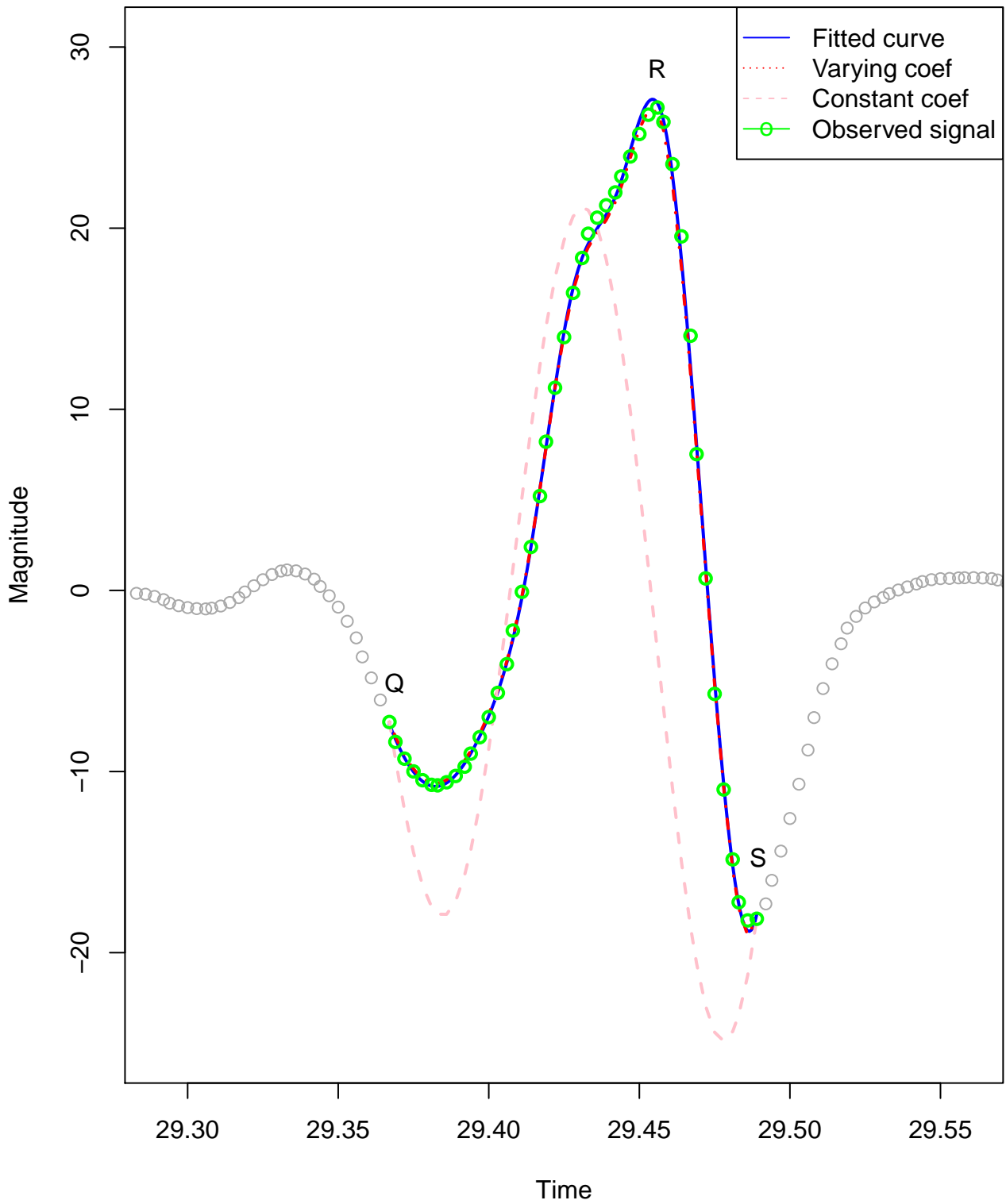

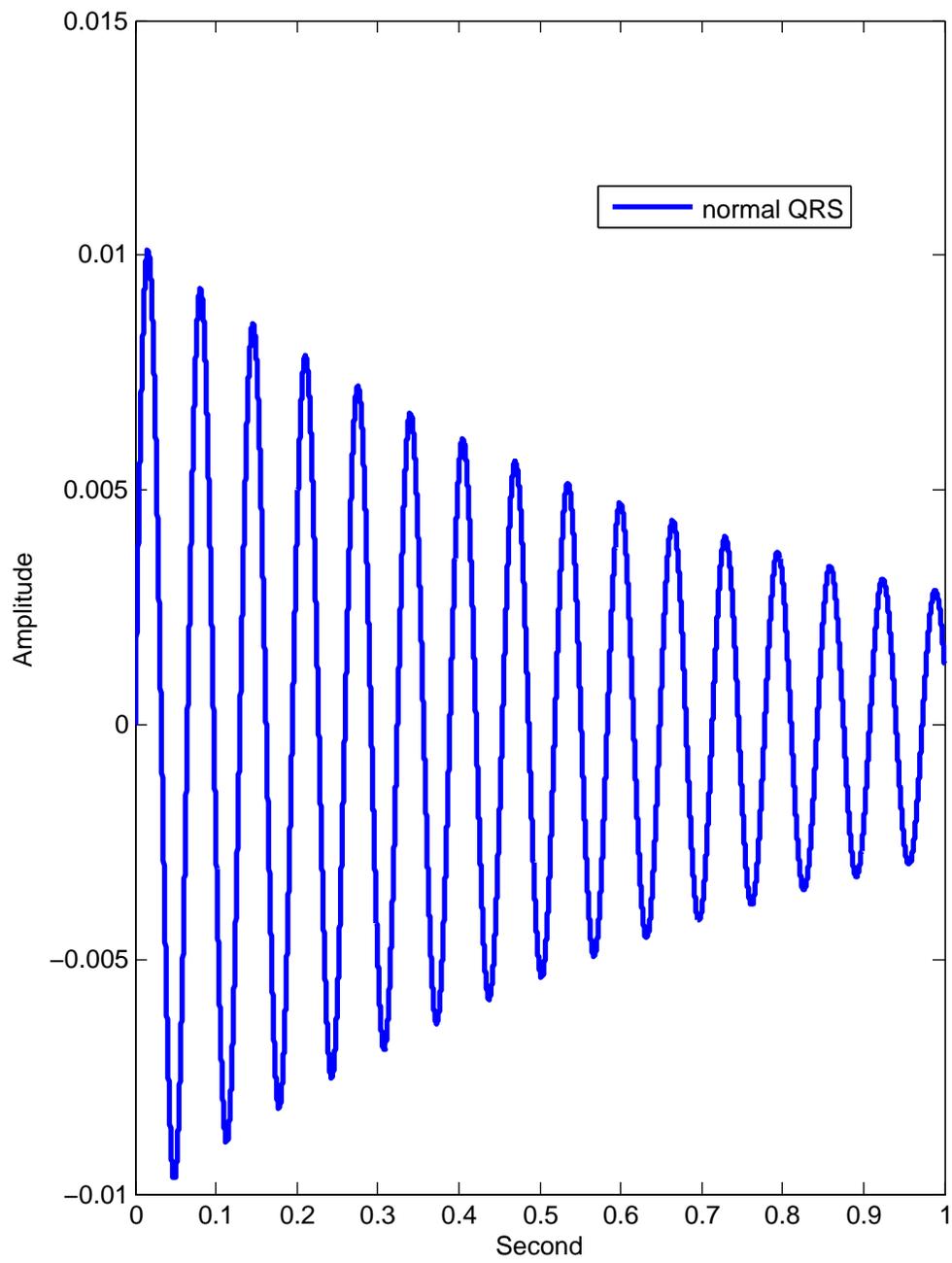

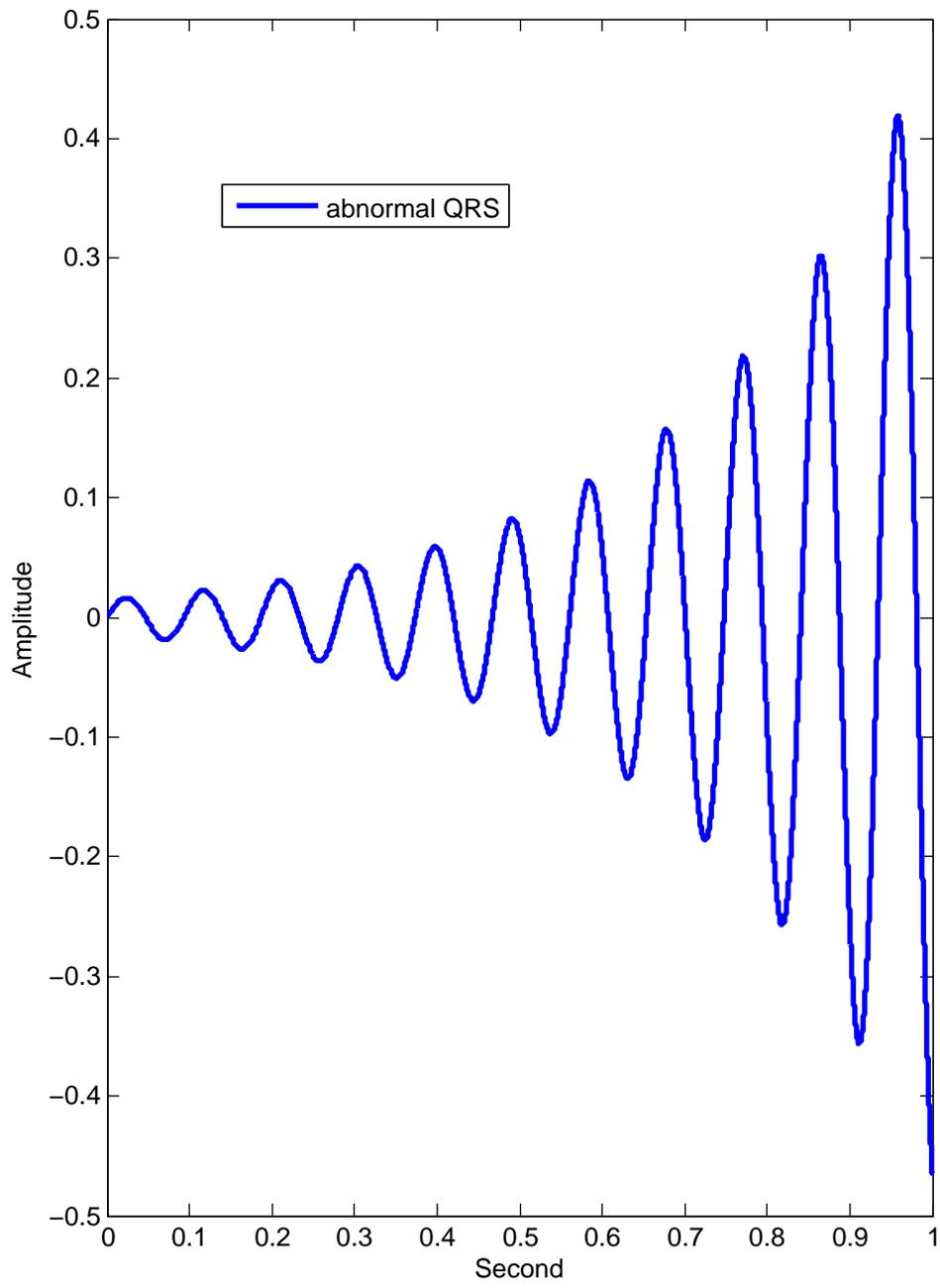

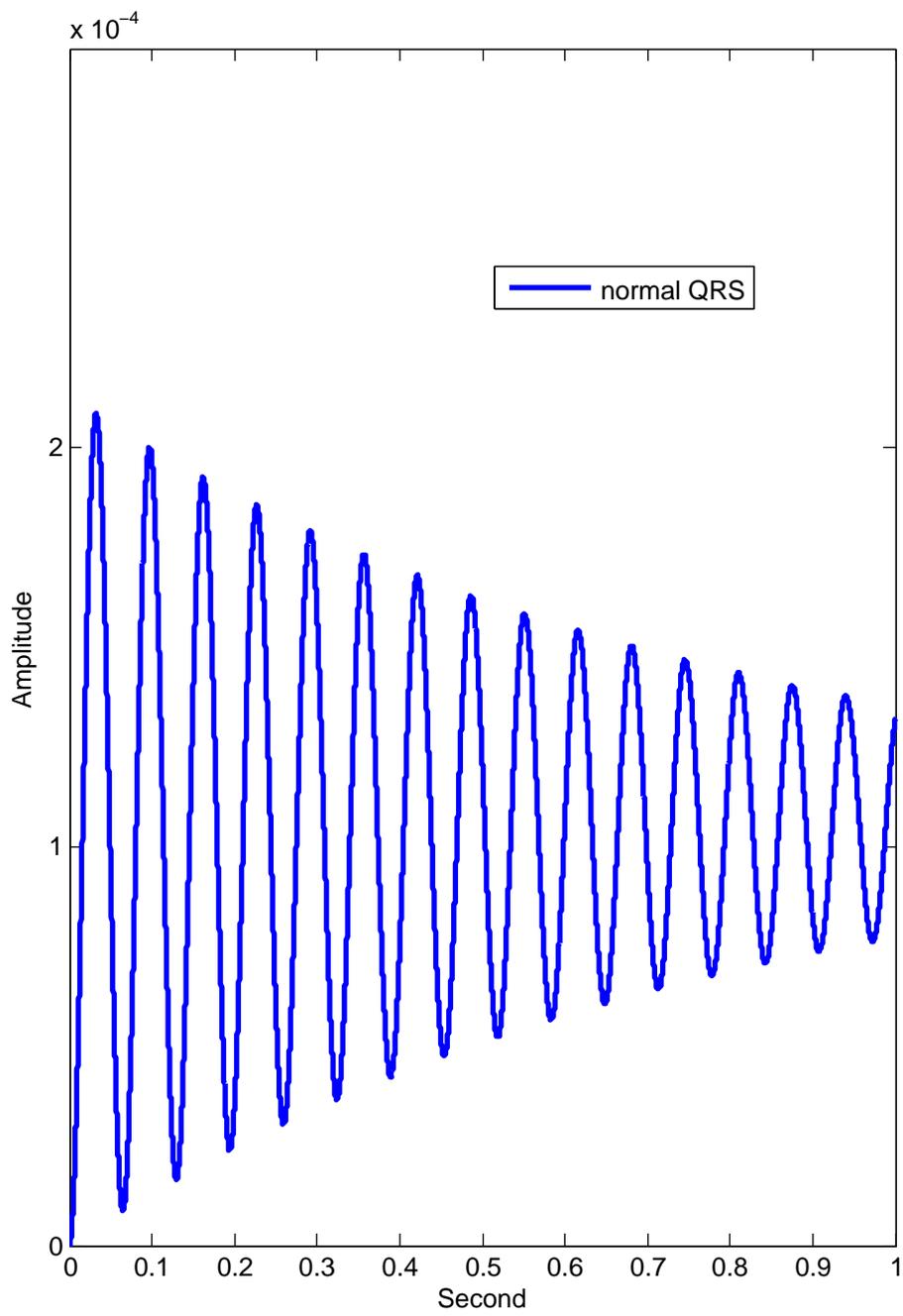

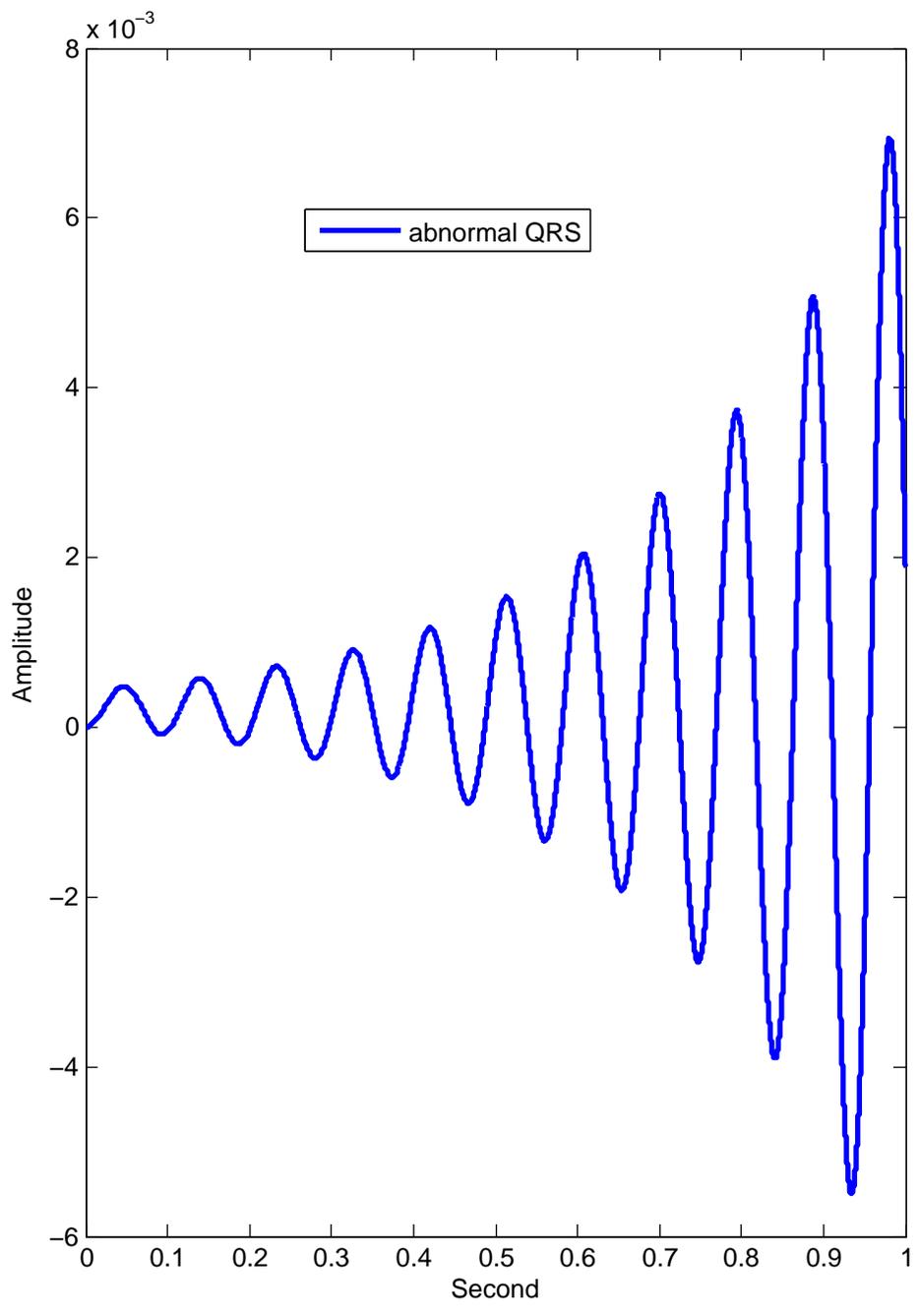